\documentclass[10pt,twocolumn,letterpaper]{article}
\usepackage[pagenumbers]{wacv} 

\usepackage{graphicx}
\usepackage{amsmath}
\usepackage{amssymb}
\usepackage{booktabs}
\usepackage{graphicx}
\usepackage{tabularray}
\usepackage{diagbox}
\usepackage{booktabs}
\usepackage{multirow}
\usepackage{caption}
\usepackage{subcaption}
\usepackage{tabularx}
\usepackage{array}
\usepackage{numprint}
\usepackage{float}
\newcolumntype{Y}{>{\centering\arraybackslash}X}

\usepackage[pagebackref,breaklinks,colorlinks]{hyperref}

\newcommand{\zhat}{\hat{\textbf{z}}_{t'}}
\newcommand{\znormal}{\textbf{z}_t}

\usepackage[capitalize]{cleveref}
\crefname{section}{Sec.}{Secs.}
\Crefname{section}{Section}{Sections}
\Crefname{table}{Table}{Tables}
\crefname{table}{Tab.}{Tabs.}

\begin{document}

\title{Uncovering Hidden Subspaces in Video Diffusion Models Using Re-Identification 

}

\author{Mischa Dombrowski$^1$ \qquad Hadrien Reynaud$^2$ \qquad Bernhard Kainz$^{1,2}$\\
$^1$Friedrich--Alexander--Universit\"at Erlangen--N\"urnberg \qquad
$^2$Imperial College London
\\
{\tt\small mischa.dombrowski@fau.de}}
\maketitle

\begin{abstract}
Latent Video Diffusion Models can easily deceive casual observers and domain experts alike thanks to the produced image quality and temporal consistency.  
Beyond entertainment, this creates opportunities around safe data sharing of fully synthetic datasets, which are crucial in healthcare, as well as other domains relying on sensitive personal information. However, privacy concerns with this approach have not fully been addressed yet, and models trained on synthetic data for specific downstream tasks still perform worse than those trained on real data. This discrepancy may be partly due to the sampling space being a subspace of the training videos, effectively reducing the training data size for downstream models. Additionally, the reduced temporal consistency when generating long videos could be a contributing factor. 

In this paper, we first show that training privacy preserving models in latent space is computationally more efficient and generalize better. Furthermore, to investigate downstream degradation factors, we propose to use a re-identification model, previously employed as a privacy preservation filter.
We demonstrate that it is sufficient to train this model on the latent space of the video generator. Subsequently, we use these models to evaluate the subspace covered by synthetic video datasets and thus introduce a new way to measure the faithfulness of generative machine learning models.
We focus on a specific application in healthcare -- echocardiography -- to illustrate the effectiveness of our novel methods.  
Our findings indicate that only up to 30.8\% of the training videos are learned in latent video diffusion models, which could explain the lack of performance when training downstream tasks on synthetic data. 
\end{abstract}
\section{Introduction}
\begin{figure}[h!]
    \centering
    \includegraphics[width=\linewidth]{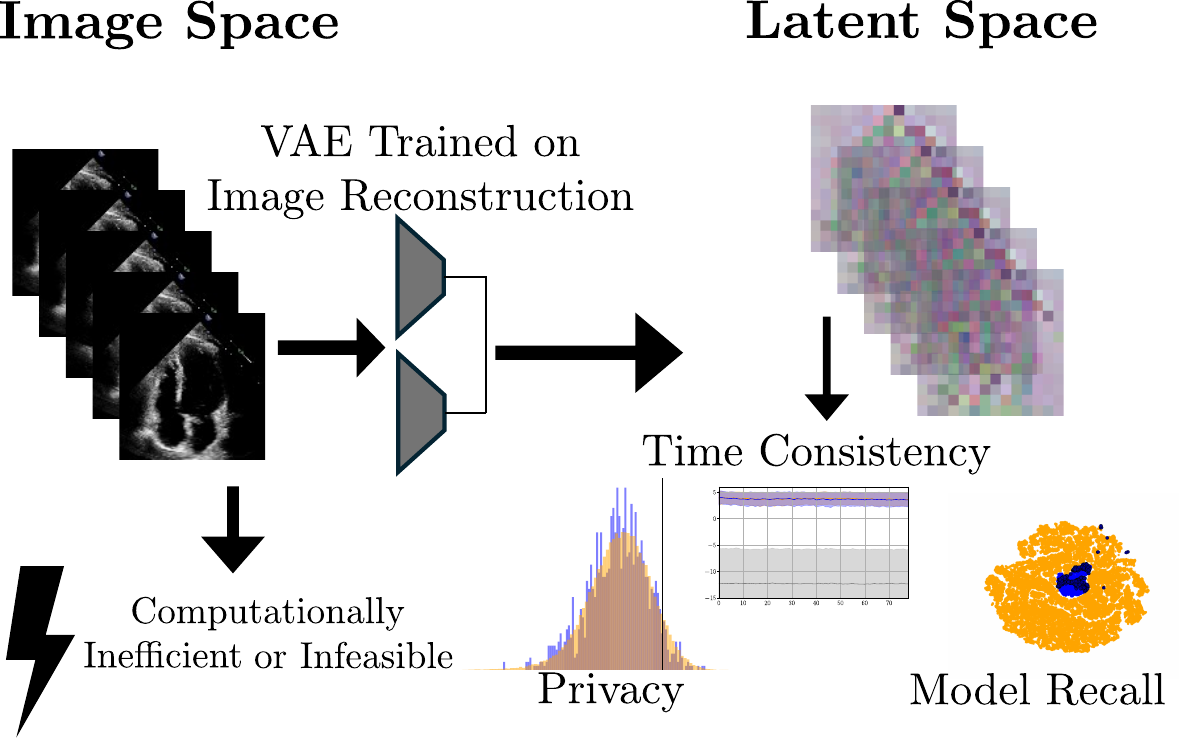}
    \caption{We propose leveraging Variational Autoencoders (VAEs), which were trained to enable the training of video diffusion models, to enhance the computational efficiency of models used for privacy filtering. Additionally, we demonstrate how the trained filter model can be applied to various other tasks, such as evaluating temporal consistency and model recall.}
    \label{fig:abstract}
\end{figure}

Latent Video Diffusion Models became popular due to their realistic depiction of synthetic scenes, mostly framed as text-to-video tasks~\cite{blattmann2023stable}. However, video synthesis also has important applications in domains beyond entertainment, such as healthcare. Analysis of diagnostic counterfactuals~\cite{reynaud2022d}, faithful representation of anatomy~\cite{qiao2023cheart}, automated diagnosis~\cite{muller2023multimodal} or data enrichment and domain generalization~\cite{ktena2024generative,kazerouni2023diffusion} are only a few examples. Images and videos are the most prevalent data structure for point-of-care and clinical diagnosis of disease with medical ultrasound imaging. The most common applications are prenatal screening and echocardiography. Echocardiography is of crucial importance to prevent avoidable deaths caused by cardiac disease.  
Ejection fraction is one of the main diagnostic measurement in echocardiography due to its correlation to ventricular function assessment \cite{reddy2023video}. It describes the ratio of change in cardiac volume during different phases of the cardiac cycle \cite{ouyang2020video}. For example, it serves as an important marker for chemotherapy dosing in pediatric ultrasound \cite{LEERINK202162} and pacemaker placement \cite{shah2021}. Consequently, interest in the automatic assessment of ejection fraction from videos has grown \cite{ouyang2020video,reynaud2021ultrasound,dai2022cyclical}. Similar to other medical fields, these models suffer from challenges related to difficult data acquisition processes, as well as robustness and domain generalization problems \cite{ouyang2020video}. 

One potential solution is the use of generative models to learn and generate videos from these domains, with the added potential to publish the data with privacy guarantees \cite{dombrowski2023quantifying,reynaud2024echonet}. This option got traction with recent advances in the quality of video generation, which predominantly operate in a latent space \cite{an2023latent,blattmann2023stable,blattmann2023align,ho2022imagen,reynaud2024echonet,khachatryan2023text2video}. 

To publicly release these synthetic videos, we need privacy measures to ensure that the synthetically generated videos are not memorized from the training data, which could lead to privacy concerns. While current approaches have begun to apply these techniques to ensure privacy \cite{dar2024unconditional,reynaud2024echonet}, there has been no evidence that we can directly ensure privacy in the latent space of diffusion models.

\noindent\textbf{Contribution:} Our contributions can be divided into two parts. First, we demonstrate that training privacy models in latent space is computationally more efficient and results in better generalization. This is illustrated in \cref{fig:abstract}. Secondly, we are the first to use a model trained for privacy to evaluate the learned subspace of the training dataset and the temporal consistency of the generated videos. In summary, we achieve the following:

\begin{enumerate}
    \setlength\itemsep{5pt}
    \parskip0pt
    \item We are the first to evaluate the feasibility of privacy-preserving models on synthetic videos, providing evidence that privatizing the latent model directly is sufficient to ensure the privacy of the entire video model.
    \item We show that training identification models in the latent space is computationally more effective than training them in image space. This approach also enhances the robustness of the learned representation of the privacy model, improving its generalization potential to new datasets.
    \item We establish a privacy filter that is simple to interpret and computationally efficient. We find that using the prediction head directly works better than taking the correlation coefficient.
    \item We illustrate how trained privacy models can be used to evaluate several other quality measurements, including the temporal consistency of diffusion-generated videos and the completeness of the generated subspace.
    \item We reveal that current generative models tend to only learn a distinct subspace of the training set, which might be the main reason for reduced performance of downstream models trained on synthetic data.
\end{enumerate}

\section{Related Work}
\noindent\textbf{Latent Generative Models:} Recent advancements in diffusion models have significantly increased interest in image and video synthesis. 
Key improvements, such as enhancements in the noise schedule, the learning target, and the sampling procedure, have contributed to their success \cite{ho2020denoising,song2020denoising,dhariwal2021diffusion}. 
The most substantial leap was the idea of performing image generation in latent space \cite{rombach2022high}. 
Instead of generating images directly, models are trained to generate image embeddings that can be decoded by a separate model. 
Despite the added complexity of this two-stage process, the improvements in computational efficiency and feasibility have made latent generative models the de facto standard for image and video generation.
\cite{pinaya2022brain} leveraged these advancements to train a 3D generative model and published a synthetic dataset of brain images. They used a dataset of \numprint{31740} images to train and sample a conditional diffusion model designed to generate synthetic images conditioned on different physiological variables. They were the first to demonstrate the potential of large-scale synthetic datasharing which raised concerns about the effects on patient-privacy. 

\noindent\textbf{Privacy:} There are two main approaches currently used to ensure privacy. The first is to guarantee the privacy of the model itself, while the second ensures that the published data is private. Guaranteeing model privacy is typically achieved using differential privacy, which mathematically guarantees privacy by altering the input data or the optimization process during training \cite{dockhorn2022differentially}. The advantage is that it allows for a model budget, which can be used to trade-off between performance and privacy guarantees. However, this trade-off is often difficult to interpret, and the drop in performance can be too significant for the model to be practical. 
Other approaches include ensuring a low likelihood of models producing unique features \cite{dombrowski2023quantifying} or enhancing the generalization capabilities of diffusion models, which are closely related to memorization \cite{li2024generalization}. 

The alternative approach is to guarantee that the produced dataset is privacy-compliant, which is more straightforward as the dataset, and hence the search space for privacy concerns, is inherently limited \cite{fernandez2023privacy,reynaud2024echonet}. The most common approach to assess privacy is to use a re-identification model, which is trained to predict whether two scans come from the same person or not \cite{packhauser2022deep,fernandez2023privacy,reynaud2024echonet,dar2024unconditional}. Although this approach does not account for the need for training images in re-identification models, it provides very strong privacy guarantees. Therefore, we will follow the same approach to validate the privacy of our generated videos. 

\noindent\textbf{Training on Synthetic Data:} 
There is substantial evidence that generative data is still not as effective as real data. One example is the recursive application of diffusion models on their own datasets, as shown in \cite{shumailov2023curse,alemohammad2023self}. These applications demonstrate that while the generated datasets are of high sample quality, they still contain hidden features that successive models will pick up and amplify. \cite{alemohammad2023self} show that recursively applying these models leads to an increasing mode shift and reduced diversity.

Another key piece of evidence is that models trained on synthetic data currently do not achieve the same performance levels on specific downstream tasks, such as classification and segmentation, as models trained on real data. The benchmark for fully synthetic datasets remains their real counterparts \cite{reynaud2024echonet,fernandez2022can,fernandez2023privacy}. This aligns with the original motivation of generative models to augment and impute existing datasets, but it currently hinders the feasibility of using these models for data sharing.

We hypothesize that part of the reason for this performance gap is that the methods currently used for quantifying generative models do not successfully capture all the relevant properties necessary to outperform real datasets. This is in line with concurrent work that critisizes and tries to improve on metrics used to evaluate generative models \cite{kynkaanniemi2019improved,stein2024exposing,dombrowski2023quantifying}. A more thorough analysis of the generated datasets is necessary to bridge this gap.

\section{Method}
\begin{figure*}[t]
    \centering
    \includegraphics[width=\linewidth]{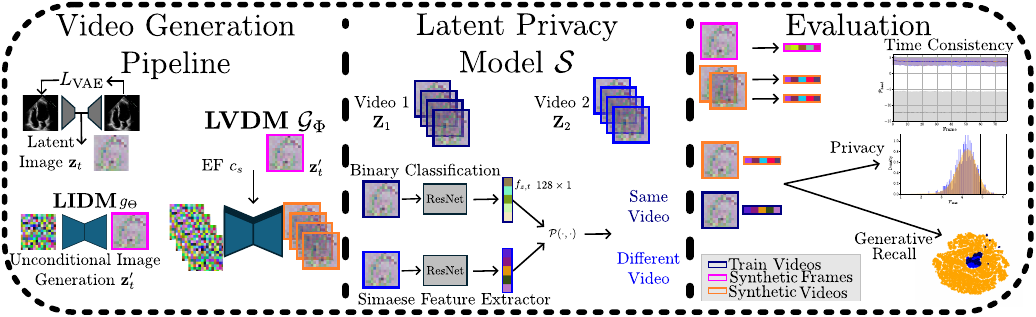}
    \caption{Overview of our approach: We take the LIDM and LVDM from~\cite{reynaud2024echonet} (left) for video generation. Our latent privacy model is based on~\cite{packhauser2022deep}. We show, that privacy regularization methods are more reliable in latent than in image space. Then we use the privacy model to evaluate temporal consistency, and generative model recall, which is a measure of how many of the training images are learned by the model without raising privacy issues.}
    \label{fig:Overview}
\end{figure*}

To generate synthetic datasets while preserving privacy, we first train a generative model $ \mathcal{G}_{\Phi}(c_s) $ to learn the data distribution $ p_{data}(\mathbf{X}|c_s) $ and subsequently sample from this model. $c_s$ is the variable that we want to learn to predict from a purely synthetic dataset.  
The process begins with a dataset $ \mathcal{X} $ consisting of real videos $ \mathbf{X} \in \mathbb{R}^{l \times c \times h \times w} $. 
We split this dataset into two disjoint proper subsets $\mathcal{X}_{\text{train}}$ and $\mathcal{X}_{\text{test}}$.
The model $ S_{\Phi}(c_s) $ is then trained to produce a synthetic dataset $ \mathcal{X}_{syn} $. 
To ensure privacy, we apply a privacy filter $\mathcal{S}$ to obtain an anonymized dataset $ \mathcal{D}_{ano} $.
For the downstream task, we aim to predict $ p(c_s|\mathbf{X}) $ and evaluate the model's performance on real data.

Current data generation methodologies predominantly operate in latent space \cite{rombach2022high, reynaud2024echonet}. 
This offers several advantages: it allows quicker training, lowers computational requirements, enables faster sampling, requires less data, and the two-stage approach allows for information compression. 
Consequently, the generative model can concentrate on learning the most relevant information.
Therefore, we take the variational autoencoder (VAE) from \cite{reynaud2024echonet} which is based on \cite{rombach2022high}. 
The VAE is trained on the task of image reconstruction. This means that we split the videos into frames $\mathbf{x}_t$ where t denotes the frame number $t \in \{1, \dots, l\}$. The architecture consists of an encoder $Enc$ and a decoder $Dec$. 
The purpose of the encoder is to compress the input into a bottleneck latent representation $\textbf{z}_t$ which can then be used as input to the decoder to reconstruct the original frame $\Tilde{\textbf{x}_t}$, \emph{i.e.}, $\Tilde{\textbf{x}_t} = Dec(Enc(\textbf{x}_t)) = Dec(\textbf{z}_t)$. The latent representation $\textbf{z}_{t}$ has three downsampling layers, which means that it is only $\frac{1}{8}$ of the size in each physical dimension and has a channel size of four, which is equal to a total compression factor of 48. 
Each latent feature consists of a mean and a variance, thus they represent a Gaussian distribution from which we can sample. 

The VAE is optimized to retain perceptual quality. First, we employ two reconstruction-based losses, which are a standard $L_1$ loss and LPIPS \cite{zhang2018unreasonable}, with a patch-wise learned feature extraction-based loss. 
To retain a small latent space, a low weighted Kullback-Leibler loss between $z_{t}$ and a standard normal distribution for regularization is applied.
Additionally, an adversarial loss from a patch-based \cite{isola2017image} discriminator $\mathcal{R}_\psi$ is trained on distinguishing between real and reconstructed images\cite{dosovitskiy2016generating,esser2021taming}.
In summary, this leads to:
\begin{multline}
    L_{\text{VAE}} = \min_{Enc, Dec} \max_{\psi}(L_{\text{rec}}(\textbf{x}_t, \Tilde{\textbf{x}_t}) - L_{\text{adv}}(\Tilde{\textbf{x}_t}) +\\  \text{log}\mathcal{R}_\psi (\textbf{x}_t, \Tilde{\textbf{x}_t}) + L_{\text{reg}}(\textbf{z}_t).
\end{multline}
The latent representation enables quicker training and sampling from the diffusion model, which is trained on the latent videos $\textbf{Z} = Enc(\textbf{X})$ encoded frame-by-frame by the VAE. 

\noindent\textbf{Generative Model:} We train, sample and use the same architecture for the diffusion models as discussed in~\cite{reynaud2024echonet}, which describes the state-of-the-art for generating synthetic medical ultrasound videos. Importantly, our generative models work entirely in the latent space, \emph{i.e.}, they are trained on latent videos $\textbf{Z}$ and produce synthetic latent video $\textbf{Z'}$.
The architecture consists of two parts: a latent image diffusion model (LIDM) and a latent video diffusion model (LVDM). The LIDM  $g_{\Theta} $ is an unconditional latent diffusion model trained on single frames from videos $\textbf{z}_t$ to generate synthetic frames $\textbf{z}_t'$. The goal is to use them as conditioning for the anatomy of the synthetic videos. 
The LVDM $\mathcal{G}_{\Phi}(c_s, \textbf{z}_t') $ is conditioned on a synthetic conditioning frame $\textbf{z}_t'$ and a regression value $c_s$, which in our case is an ejection fraction (EF) score: a standard parameter for the systolic function of the heart \cite{reddy2023video}.
From these synthetic videos, we can train a downstream model to predict EF and test on videos from real hearts. 

\noindent\textbf{Latent Privacy Model:} \label{sect:LatentPrivacyModel}
Since we are working on videos, unlike existing privacy methods \cite{packhauser2022deep, dar2024unconditional}, we do not rely on augmentation to learn meaningful representations to privatize our data. 
Instead, we can take different frames from the same video as augmentations to train a self-supervised feature extractor, that will learn to differentiate between different anatomies.
As backbone, we follow the architecture proposed by \cite{packhauser2022deep} to train a siamese neural network model $\mathcal{S}(\znormal, \zhat)$ for binary classification of whether the latents $\znormal$ and $\zhat$ come from the same video.   
The feature encoding part of the architecture, previously referred to as our filter $\mathcal{F}$, is a ResNet-50 \cite{he2015deep} pre-trained on Imagenet \cite{deng2009imagenet}. 
This feature encoder $\mathcal{F}$ computes the feature representation $f_{z, t}$ of each latent input frame $\znormal$.
The final prediction is as follows: 
\begin{multline}
    \mathcal{S}(\textbf{z}_t, \hat{\textbf{z}}_{t'}) = 
    \sigma(\text{MLP}(|\mathcal{F}(\znormal) - \mathcal{F}(\zhat)|)) = \\
    \mathcal{P}(\mathcal{F}(\znormal), \mathcal{F}(\zhat)) = 
    \mathcal{P}(f_{z, t}, f_{\hat{z}, t'}),
    \label{eq:featuremodel}
\end{multline}
where $\mathcal{P}$ can be seen as the predictor function that considers the learned features and predicts whether the frames come from the same or from a different video.

To compare different approaches and analyze the interpretability of the feature space extracted by $\mathcal{F}$ we experiment with multiple distance metrics. These metrics can be interchangeably used as distance metrics or predictor function $\mathcal{P}$. The default (Pred) is the classification head learned according to Eq.~\ref{eq:featuremodel}. 
We also consider the correlation coefficient according to~\cite{dar2024unconditional}. 
\cite{dar2024unconditional} use an additional contrastive learning technique to increase the distance for negative pairs. 
We skip this step as the smaller changes coming from using consecutive frames instead of image augmentations, such as flipping, trivially leads to positive pairs being mapped near the zero element. 
The whole pipeline is illustrated in \ref{fig:Overview}. 

\section{Experiments}
\label{sec:experiments}
\noindent\textbf{Datasets:}
EchoNet-Dynamic is a dataset consisting of \numprint{10030} ultrasound videos from unique patients \cite{ouyang2020video}. We apply the official data splits with \numprint{7465} patients for training, \numprint{1277} for validation and \numprint{1288} for testing. For each video, we have a ground-truth ejection fraction value available. 
The videos are standardized to a resolution of $112 \times 112$ pixels, and have a variable number of frame.
EchoNet-Pediatric is an ultrasound video dataset, specifically built from scans of younger patients ranging from 0 to 18 years old. The dataset is split into apical 4-chamber (A4C) and parasternal short-axis (PSAX) views \cite{reddy2023video}. 

\noindent\textbf{Metrics:}
To assess the feasibility of measuring temporal consistency, we compute the mean correlation coefficient (MCC) according to:
\begin{equation}
    \label{eq:mcc}
    \text{MCC} = \sum_{z \in \mathcal{D}_{\text{test}}} \sum_{t \in \{1,\dots,l\}} \sum_{t' \in \{1,\dots,l\}} \text{corr}(f_{z,t}, f_{z, t'}), 
\end{equation}
where corr is the Pearson's correlation coefficient. 
To assess the performance of the privacy model, we compute the area under the receiver operating characteristic curve (AUC-ROC). 
To compute the accuracy of the regression model for the downstream task, we compute the mean absolute error (MAE), the root mean squared error (RMSE) and the coefficient of determination $R^2$. 

\subsection{Superiority of Latent Privacy Models}
\label{sec:golatent}
First, we compare latent and image space privacy models. We show that both are equally good on a training set, but the latent model demonstrates better generalization capabilities. 
For each of the three datasets (EchoNet-Dynamic, EchoNet-Pediatric A4C and EchoNet-Pediatric PSAX), we train two privacy models $\mathcal{S}$. One is trained directly on the videos $\textbf{X}$ from the training set $\mathcal{X}_\text{train}$ and the second one is trained on the latent videos \textbf{Z} using $\mathcal{E}$. 
We train the image space models for \numprint{1000} epochs with a batch size of 128 and an early stopping set at 50 epochs, following to \cite{packhauser2022deep}. The latent space models run with a larger batch size of \numprint{1024}, a higher learning rate of 5e-4, and a more patient early stopping set to 150 epochs. 
For testing, we go through the test dataset and randomly select either a frame from the same video or a different video with equal probability. 

\begin{figure}[]
    \centering
    \begin{subfigure}[b]{\linewidth}
        \centering
        \includegraphics[width=1.0\linewidth]{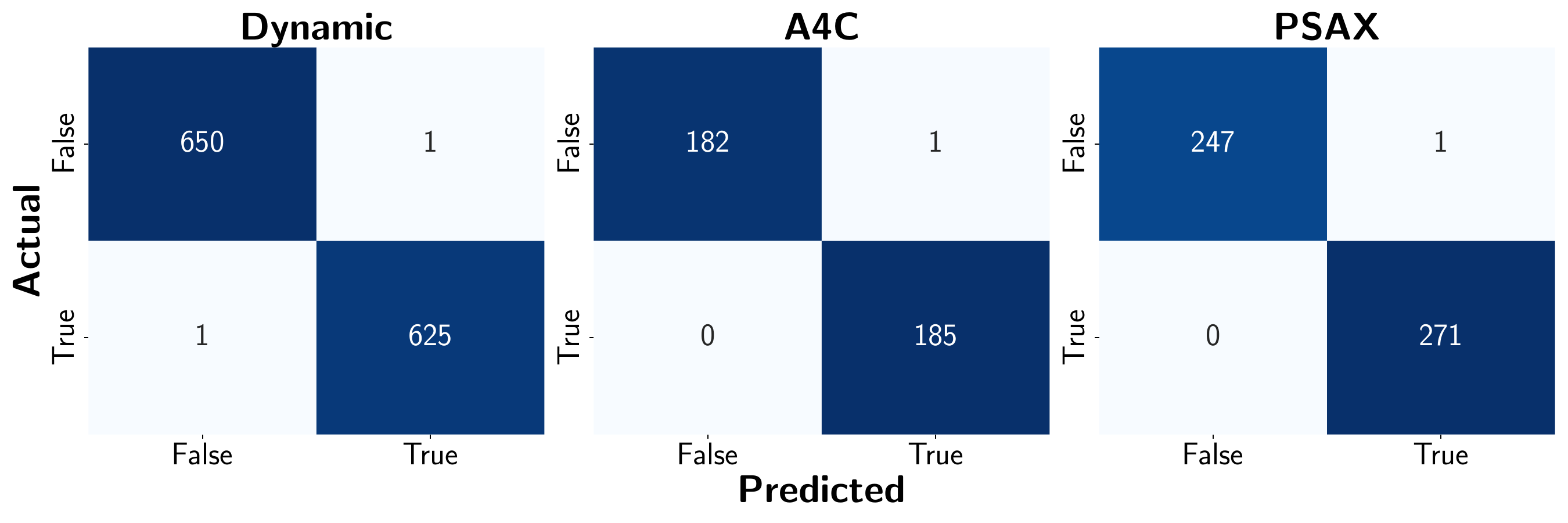}
        \caption{Image space}
    \end{subfigure}
    
    \begin{subfigure}[b]{\linewidth}
        \centering
        \includegraphics[width=1.0\linewidth]{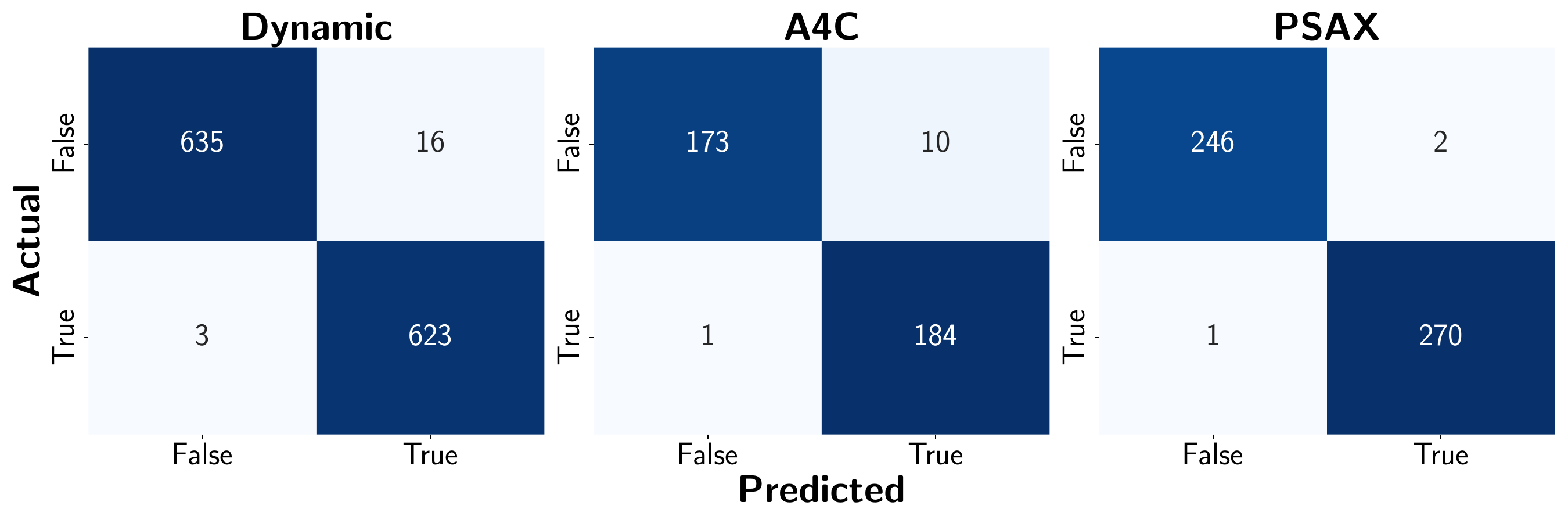}
        \caption{Latent space}
    \end{subfigure}
    \caption{Confusion matrices for the different datasets computed in image and latent space.}
    \label{fig:latent_space_confusion}
\end{figure}

\noindent\textbf{Computational improvements by using embeddings:} A single epoch takes roughly three minutes in image space, which amounts to a total of 50 hours of training. 
In latent space, training the model only takes three seconds for a single epoch, which only adds up to 50 minutes per training run and an overall computational time improvement of $60\times$. 
Furthermore, we can keep the precomputed latent representation of the image in memory, which is only roughly 1/64 of the memory consumption. 
We report the results in \cref{tab:privacyresults} and \cref{fig:latent_space_confusion}. The results show that both models obtain an almost perfect performance across all datasets. 
While the image model slightly outperforms the latent model, we argue that both fall within a margin of error and that the difference could be due negligible parameters such as initialization. Therefore, the image model practically provides no real advantage over the latent model. 
To demonstrate this, we compute the generalization accuracy of the classification models when applied on one dataset and tested on the other two. The results are shown in \cref{tab:generalization}, where we observe high test accuracies throughout all models. 
This time, however, we see that the latent models are better at generalization, which is partially due to the fact that the latent space was trained jointly on all the datasets.  
Furthermore, if we compute the MCC across the entire dataset we observe that it is notably higher in the latent space, \emph{i.e.}, the score correlation between features computed for frames of the same real videos is higher when we use the latent model. 
This higher score will also improve experiments on temporal consistency, which we will show in~\cref{sec:videoconsistency}. 
The increased interpretability is due to the stronger correlation between different frames of the same video.
Consequently, we will stay in latent space for the remainder of our experiments. 

\begin{table*}[]
\centering
\begin{tabular}{lllllll} 
\toprule
& \diagbox{Dataset}{Metric} & AUC                   & Accuracy & F1-Score & Precision & Recall  \\
\midrule

\multirow{3}{*}{\rotatebox{90}{Image}}  &Dynamic  & 1.000 [0.999 - 1.000] & 0.998 & 0.998 & 0.998 & 0.998   \\
                                        &a4c      & 1.000 [0.999 - 1.000] & 0.997 & 0.997 & 0.995 & 1.000   \\
                                        &psax     & 1.000 [0.999 - 1.000] & 0.998 & 0.998 & 0.996 & 1.000   \\

\midrule
\multirow{3}{*}{\rotatebox{90}{Latent}} &Dynamic  & 0.998 [0.996 - 1.000] & 0.992 & 0.992 & 0.987 & 0.997   \\
                                        &a4c      & 0.991 [0.982 - 0.998] & 0.970 & 0.971 & 0.968 & 0.995   \\
                                        &psax     & 0.999 [0.999 - 1.000] & 0.994 & 0.994 & 0.993 & 0.996   \\

\bottomrule
\label{tab:privacyresults}

\end{tabular}
\caption{Quantitative results of the identification model in image and latent space. The confidence intervals come from bootstrapping the sample-wise results following \cite{packhauser2022deep}.}
\end{table*}

\begin{table*}[]
\begin{tabularx}{\textwidth}{l|YYY|YYY}
                                          & \multicolumn{3}{c}{\textbf{Image Space}} & \multicolumn{3}{c}{\textbf{Latent Space}} \\
                                          \toprule
 \diagbox{Train}{Test}                    & Dynamic      & A4C    & PSAX    & Dynamic     & A4C     & PSAX     \\
                                          \midrule
\multirow{2}{*}{Dynamic}                  & \textbf{1.000} & 0.968 & 0.978 & 0.998 & \textbf{0.995} & \textbf{0.997}\\
                                          & $0.76 \pm 0.12$ & $0.71 \pm 0.15$ & $0.71 \pm 0.15$ & \textbf{0.89}\textbf{$\pm$}\textbf{0.07} & \textbf{0.88}\textbf{$\pm$}\textbf{0.06} & \textbf{0.89}\textbf{$\pm$}\textbf{0.05}\\
                                          \midrule
                   \multirow{2}{*}{A4C}   & 0.988 & \textbf{1.000} & \textbf{1.000} & \textbf{0.994} & 0.991 & \textbf{1.000}\\
                                          & $0.72 \pm 0.16$ & $0.71 \pm 0.14$ & $0.70 \pm 0.15$ & \textbf{0.81}\textbf{$\pm$}\textbf{0.08} & \textbf{0.81}\textbf{$\pm$}\textbf{0.08} & \textbf{0.81}\textbf{$\pm$}\textbf{0.09}\\
                                          \midrule
                   \multirow{2}{*}{PSAX}  & 0.934 & 0.962 & \textbf{1.000} & \textbf{0.996} & \textbf{0.984} & 0.999\\
                                          & $0.73 \pm 0.12$ & $0.59 \pm 0.21$ & $0.68 \pm 0.13$ & 
                                          \textbf{0.79}\textbf{$\pm$}\textbf{0.10} & 
                                          \textbf{0.77}\textbf{$\pm$}\textbf{0.12}& 
                                          \textbf{0.79}\textbf{$\pm$}\textbf{0.10}\\
                                          \bottomrule
\hline
\end{tabularx}
\caption{\textbf{Generalization to new datasets:} AUC and correlation coefficient mean and standard deviation for model trained and tested on different datasets. The results show that the models trained on latent space generalize better. In five out of six cases, the latent model outperforms the image space model, while having equal performance for the case of generalizing from A4C to PSAX. We confirm this observation by looking at self-consistency of the videos. According to the correlation coefficient, the consistency is higher between the frames of the same video in latent space than it is in image space for all videos, even for the ones trained and tested on the same dataset.}
\label{tab:generalization}
\end{table*}

\noindent\textbf{Distance Metrics:} Next, we determine the appropriate metric to use for evaluating privacy.  
The distance function $\mathcal{P}$ introduced in~\cref{eq:featuremodel} is interchangeable after training the privacy model. 
To compare the different approaches, we once again employ the test setup for generalization and check how well the classification works with different prediction heads. The results are shown in Tab.~\ref{tab:distances}. It is evident that all predictions heads have very high performance. 
Especially, the high $L^1$ and $L^2$ norms values show the rich and interpretable feature space of $\mathcal{F}$. 
Intuitively, these scores mean that different frames from the same videos have very similar features. Therefore, we conclude that distances we computed are meaningful, and it is worth looking at visualizations of the feature space in \cref{sec:ModelRecall}. Additionally, we see that the trained prediction head ("pred") performs best.  

\begin{table*}[]
\begin{tabularx}{\textwidth}{l|YYY|YYY|YYY|c}
\toprule
Test      & \multicolumn{3}{|c|}{\textbf{Dynamic}} & \multicolumn{3}{c}{\textbf{A4C}} & \multicolumn{3}{|c|}{\textbf{PSAX}} &  \multirow{2}{*}{\textbf{Total}}   \\
Train     & Dynamic    & A4C   & PSAX   & Dynamic  & A4C  & PSAX  & Dynamic   & A4C  & PSAX  &\\
\midrule
$L^1$ & 0.996 & 0.987 & 0.990 & \textbf{0.998} & 0.986 & \textbf{0.990} & \textbf{0.999} & 0.998 & 0.998 & 0.994 \\ 
$L^2$ & 0.995 & 0.978 & 0.986 & \textbf{0.998} & 0.986 & 0.989 & 0.998 & 0.996 & \textbf{0.999} & 0.992 \\
Corr & 0.996 & 0.978 & 0.970 & \textbf{0.998} & 0.984 & 0.984 & \textbf{0.999} & 0.997 & 0.993 & 0.989 \\
\midrule
Pred      & \textbf{0.998} &\textbf{0.994} & \textbf{0.996}  & 0.995    & \textbf{0.991}& 0.984 &  0.997   & \textbf{1.000} & \textbf{0.999} & \textbf{0.995}\\
\bottomrule
\end{tabularx}
\caption{AUC for test and generalization using different prediction heads $\mathcal{P}$.}
\label{tab:distances}
\end{table*}

\subsection{Evaluating Privacy}
\label{sec:exp_privacy}

\begin{figure*}[htbp]
  \centering
  \begin{subfigure}{0.32\textwidth}
    \centering
    \includegraphics[width=\linewidth]{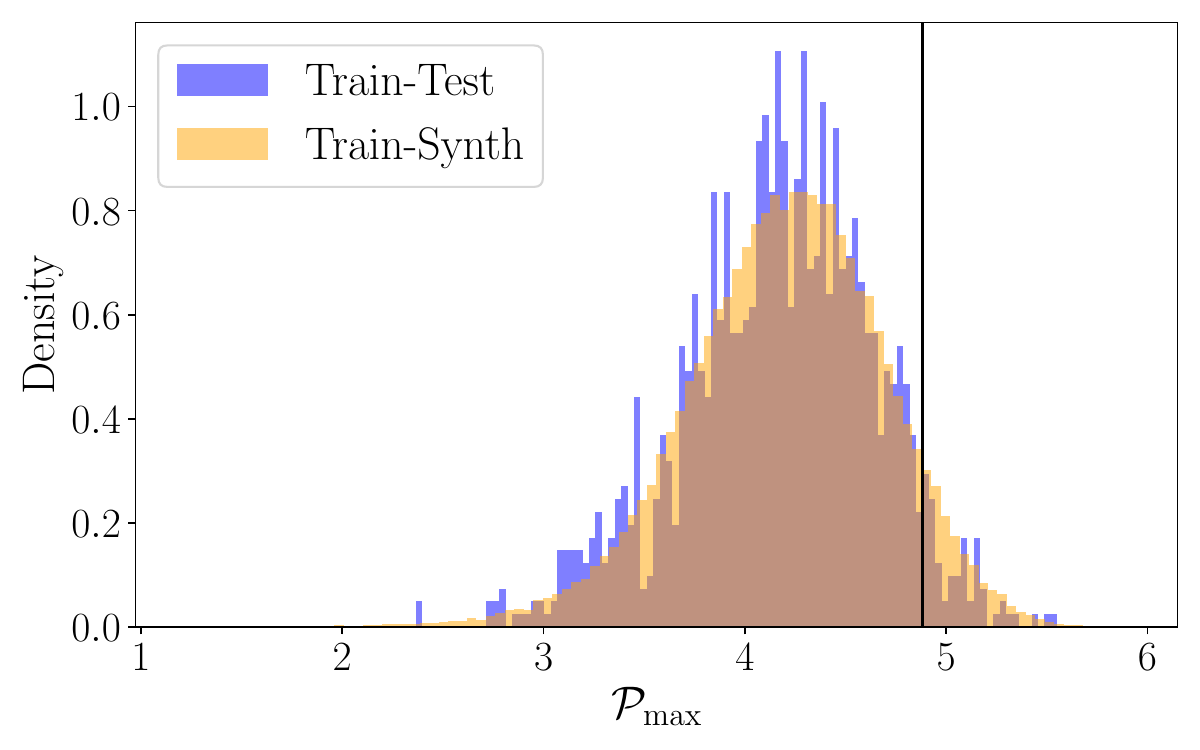}
    \caption{Dynamic}
    \label{fig:priv_Dynamic}
  \end{subfigure}
  \hfill
  \begin{subfigure}{0.32\textwidth}
    \centering
    \includegraphics[width=\linewidth]{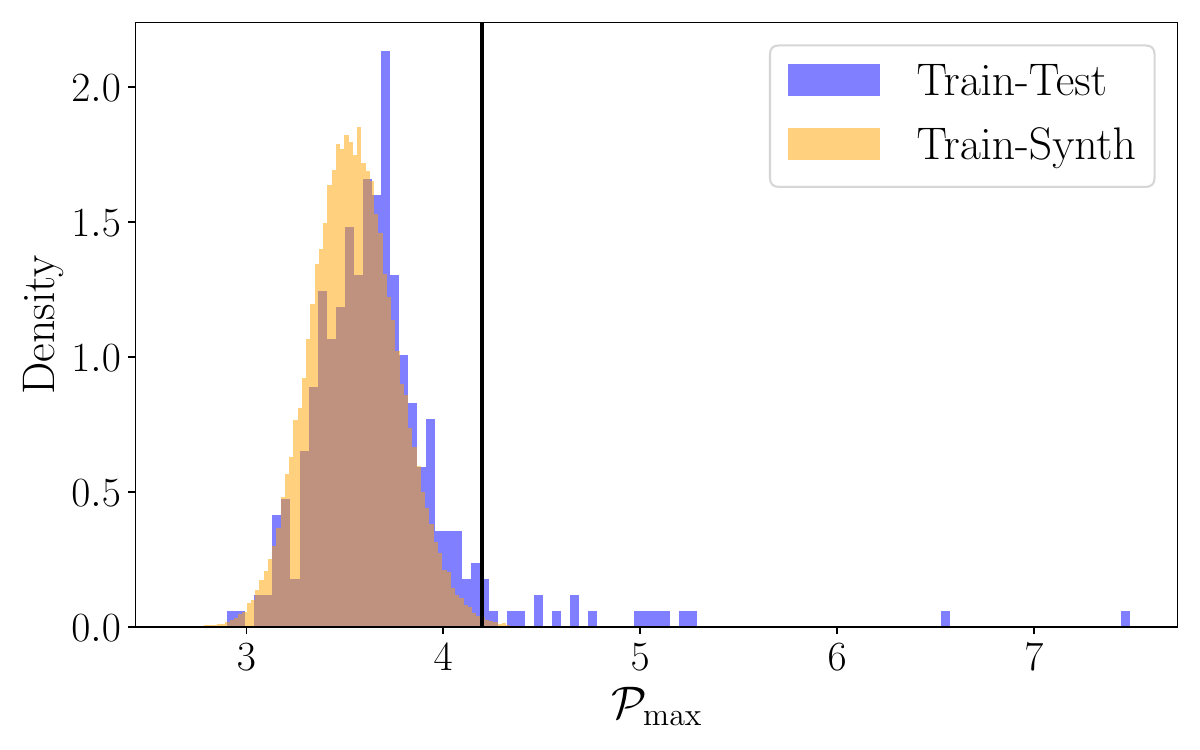}
    \caption{A4C}
    \label{fig:priv_a4c}
  \end{subfigure}
  \hfill
  \begin{subfigure}{0.32\textwidth}
    \centering
    \includegraphics[width=\linewidth]{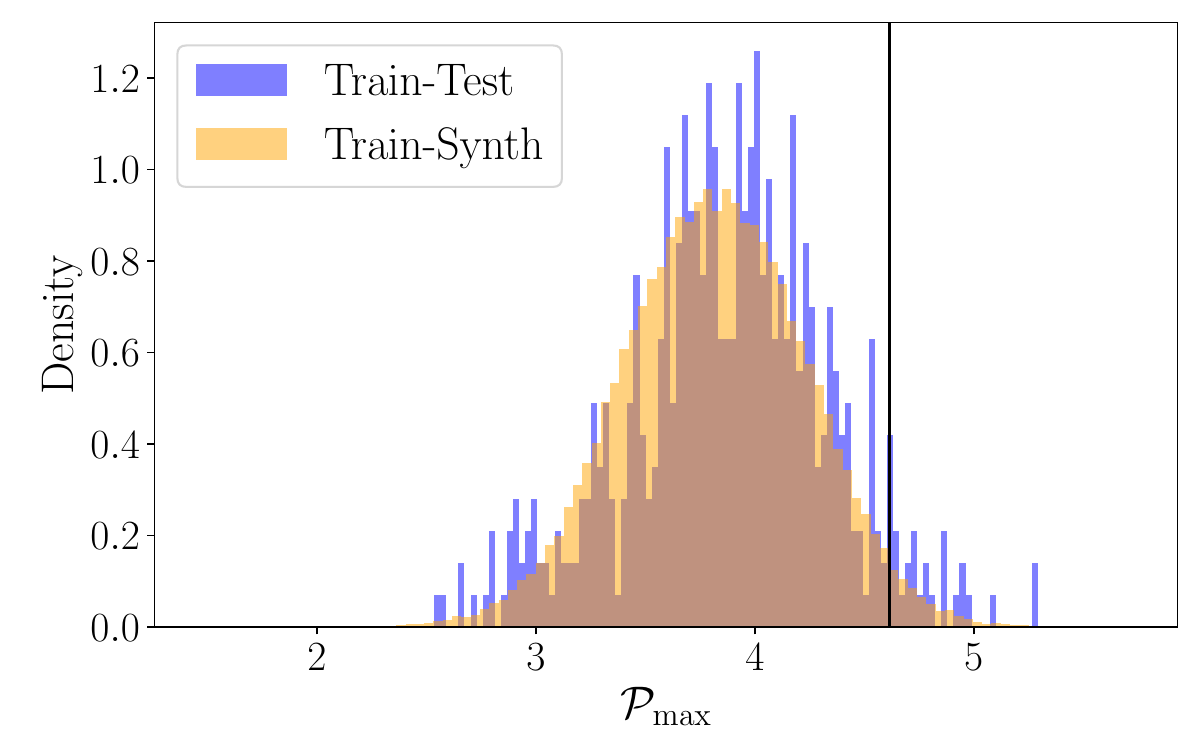}
    \caption{PSAX}
    \label{fig:priv_psax}
  \end{subfigure}
  \caption{$\mathcal{P}_{\text{max}}$ values of the distances between the training dataset and real test samples and between the training dataset and the synthetic videos. 
  The black line corresponds to the 95th percentile privacy threshold.}
  \label{fig:privacyfilter}
\end{figure*}

Following the approach proposed by \cite{dar2024unconditional} and \cite{reynaud2024echonet} we compute the privacy threshold for the videos.
We extend on the experiments conducted in \cite{reynaud2024echonet} but use the prediction head directly as a result of the superior performance reported in \cref{tab:generalization}. Additionally, we compute the mean similarity of the first frame according to the prediction head over the entire training video (\emph{i.e.}~\cref{eq:mcc} but with the "Pred" instead of "Corr"). 
Results are shown in \cref{fig:privacyfilter}. The privacy filtering works by computing $\mathcal{P}_{\text{Max}}$, which is defined as the maximum of the distance function between $\zhat$ and the first frame of each real video. 
To classify the videos as "too similar" or not, we observe the distribution of $\mathcal{P}_{\text{Max}}$ between the training and test sets. Following \cite{dar2024unconditional}, we arbitrarily chose the 95th percentile distance value as our threshold, and compute the actual distance value for each dataset using that threshold.
The privacy threshold allows us to mark all synthetic videos with a higher $\mathcal{P}_{\text{Max}}$ value as concerning in terms of dataset privacy. 
Overall, out of \numprint{100000} generated conditional images, only \numprint{7313}, \numprint{302}, and \numprint{2488} were detected as concerning and counted as memorized according to the privacy filter. 
The distribution of the highest value in the synthetic sets seems to be very close to that of the real test set, which means that the image generation model has learned to generate realistic images, without copying or memorizing its training dataset. 
The naturally occurring outliers seen, \emph{e.g.}, in \cref{fig:priv_a4c}, are not in the synthetic datasets, further manifesting our belief that the models are good in terms of privacy preservation. 
Overall, the results suggest that the video generation works well, and the synthetic videos are close to real world samples. 

\subsection{Evaluation of Generative Model Recall}
\label{sec:ModelRecall}
\begin{table}[!ht]
\resizebox{\linewidth}{!}{
\begin{tabular}{l|c|c|c}
\toprule
                      & Dynamic & A4C  & PSAX \\
                      \midrule
Train Videos          & 7465 & 2580 & 3559\\
Learned               & 1401 (18.8\%)&  794 (30.8\%) &  884 (24.8\%) \\
Memorized             & 7.31\%       &  0.3\%        &    2.5\%  \\
Learned but memorized &  5           &  3            &     2 \\
\bottomrule
\end{tabular}
}
\caption{Statistics of how many of the training videos were learned, memorized, or filtered but memorized, which means they were generated but filtered afterwards.}
\label{tab:recallsummary}
\end{table}

\begin{figure*}[htbp]
  \centering
  \begin{subfigure}{0.32\textwidth}
    \centering
    \includegraphics[width=\linewidth]{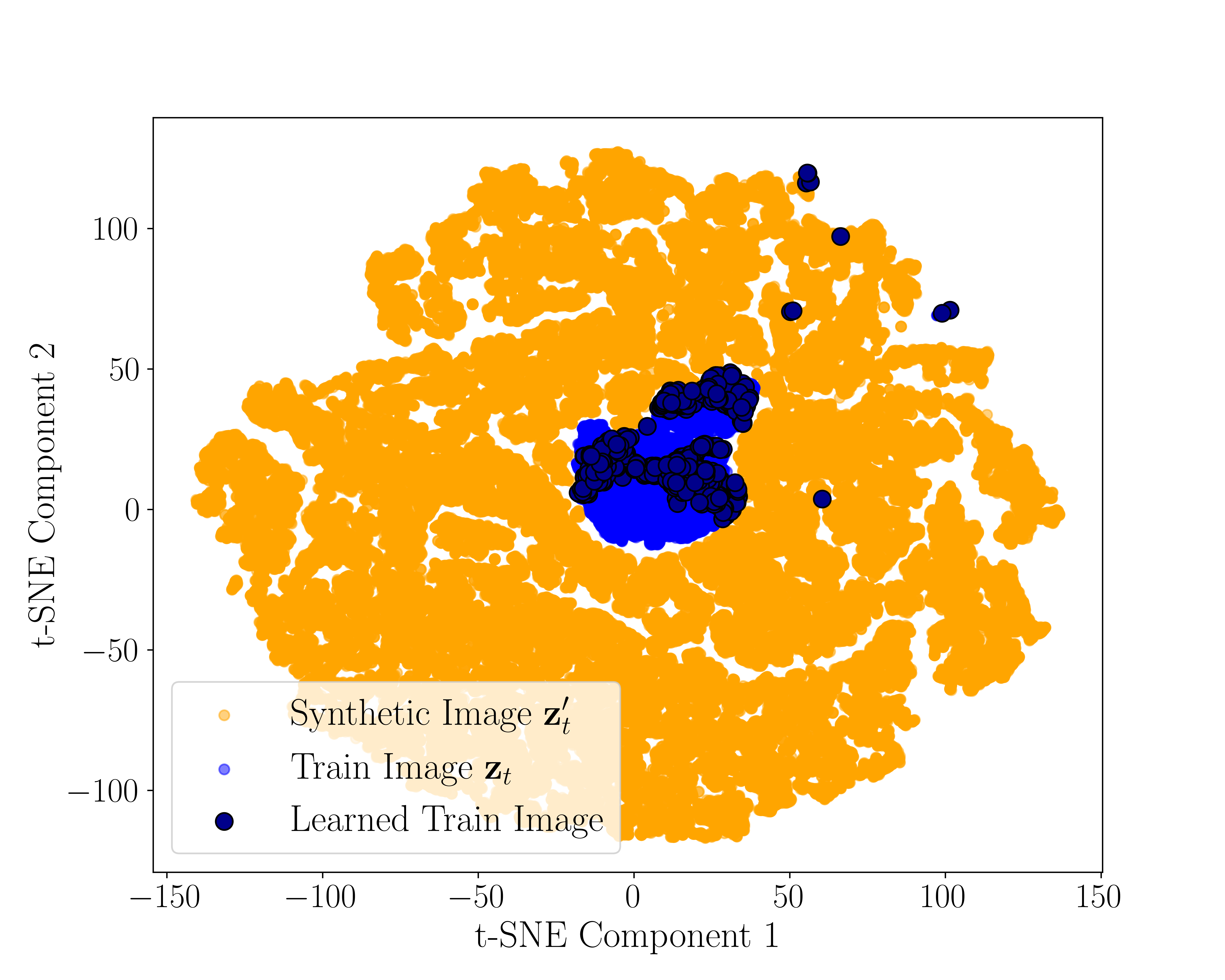}
    \caption{Dynamic}
    \label{fig:tsne_Dynamic}
  \end{subfigure}
  \hfill
  \begin{subfigure}{0.32\textwidth}
    \centering
    \includegraphics[width=\linewidth]{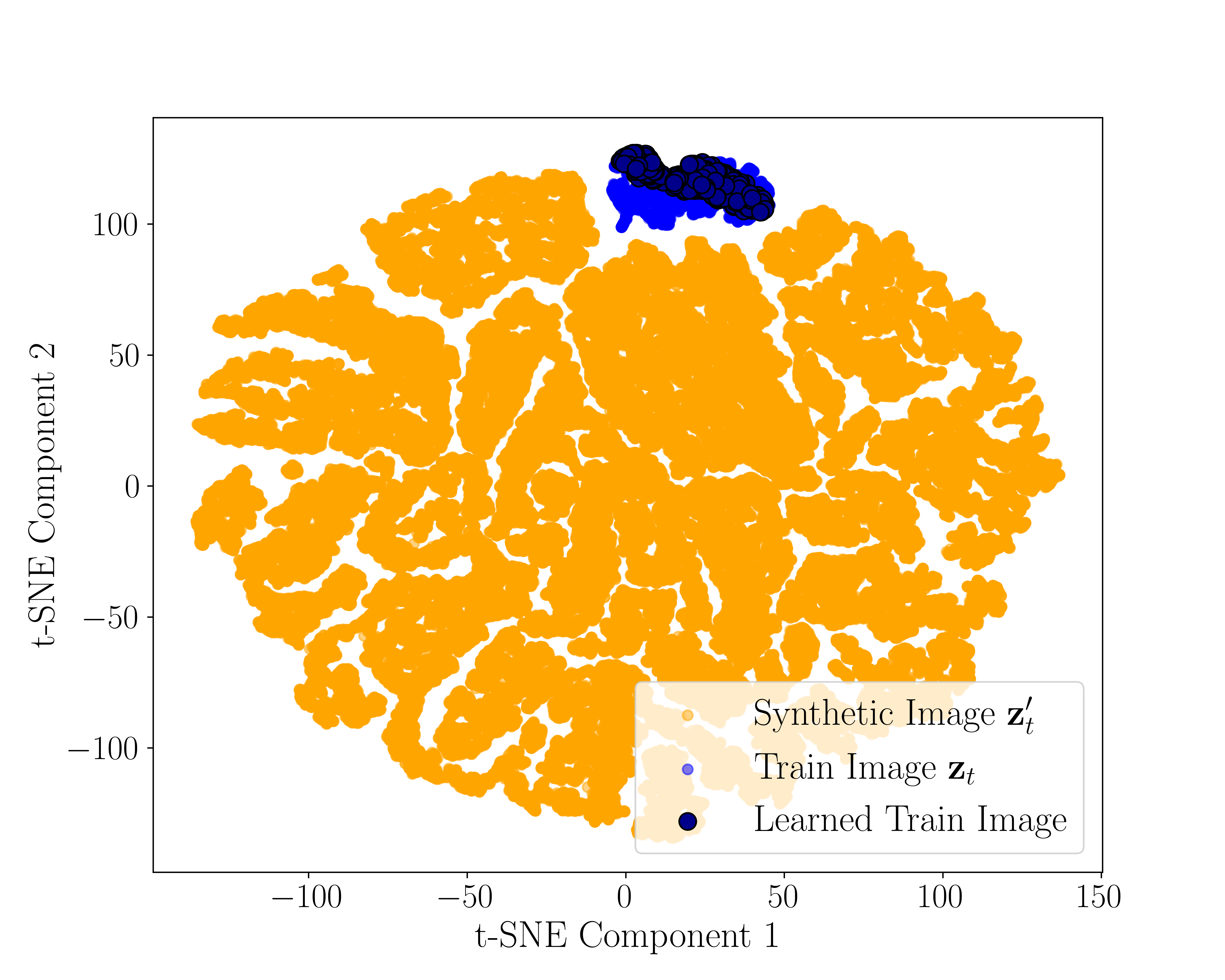}
    \caption{A4C}
    \label{fig:tsne_a4c}
  \end{subfigure}
  \hfill
  \begin{subfigure}{0.32\textwidth}
    \centering
    \includegraphics[width=\linewidth]{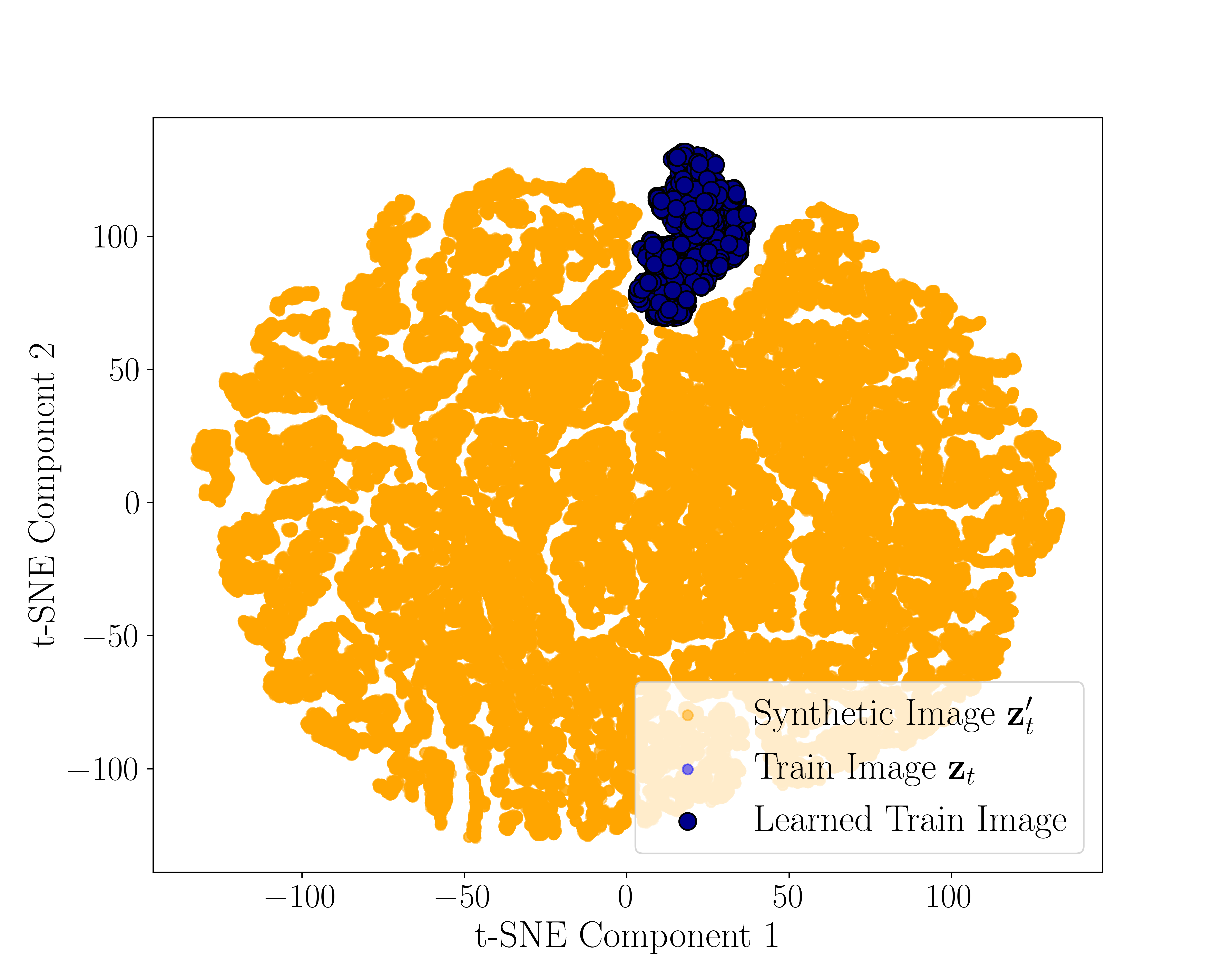}
    \caption{PSAX}
    \label{fig:tsne_psax}
  \end{subfigure}
  \caption{t-SNE plot of training and synthetic image datasets. We visualize the t-SNE components of the learned representations extracted by the privacy filter model. Synthetic samples are shown in orange and training images are shown in light blue. For every synthetic image, we apply privacy filtering. If a training image is closer to a synthetic image than the synthetic image is to all other training images, we consider the training image learned and change the color of the training image in the plot to dark blue.}
  \label{fig:tsne}
\end{figure*}

Despite the virtually infinite size of generated video datasets, downstream models trained for specific tasks on purely synthetic data are not as good as if trained on a smaller set of real training images.
We hypothesize that this is because the unconditional model only generates a subgroup of the real image distribution, similar to mode collapse for adversarial training~\cite{bang2021mggan}.
Generative model recall is the ability of the generative model to sample images that are closer to one training image than to any other training image. 
Previously we have used this definition as a filter for images that are too similar and therefore considered a privacy problem.  
Here, as a reference, we first compute this value between the train and test split of the real videos. 
For all three datasets, the percentage of test videos that have a closest neighbor in the training dataset is higher than 87\%. 
When we compute this value for the synthetic videos and the training videos, we find that in all cases it is lower than 25\%, despite the fact that we have sampled \numprint{100000} synthetic images. 
Hence, \numprint{100000} synthetic videos only represent up to 25\% of the real dataset distribution. This value can be interpreted as recall of the generative model \cite{kynkaanniemi2019improved}.
We summarize these results in \cref{tab:recallsummary}. 
We also report a category ``learned but memorized'' which means that the model was able to sample frames that are closest to a specific training sample, but they are so close that they trigger the privacy filter $\mathcal{S}$. 
Looking at the distribution of all learned images, \emph{i.e.}, how often each sample was marked as learned according to having a highest $\mathcal{P}_{\text{Max}}$ score, reveals how irregularly each sample is generated. 
For the Dynamic dataset, \numprint{50000} synthetic videos only represent 161 videos of the real dataset, 
with the most highly represented video being 802 times in the synthetic dataset according to this definition.  
Finally, we visualize this using t-SNE in \cref{fig:tsne} \cite{JMLR:v9:vandermaaten08a}. As input, we take the learned representation of the privacy model $f_{z,t}$. 
For all models, it appears that the synthetic representations are far off from the real representations, as can be seen by the fact that the synthetic cluster is much larger. 
Furthermore, we observe for Dynamic and A4C that the learned representation do not fully represent the cluster of real videos indicated by the low overlap. 

\subsection{Evaluating Video Consistency}
\label{sec:videoconsistency}
\begin{figure*}[htbp]
  \centering
  \begin{subfigure}{0.32\textwidth}
    \centering
    \includegraphics[width=\linewidth]{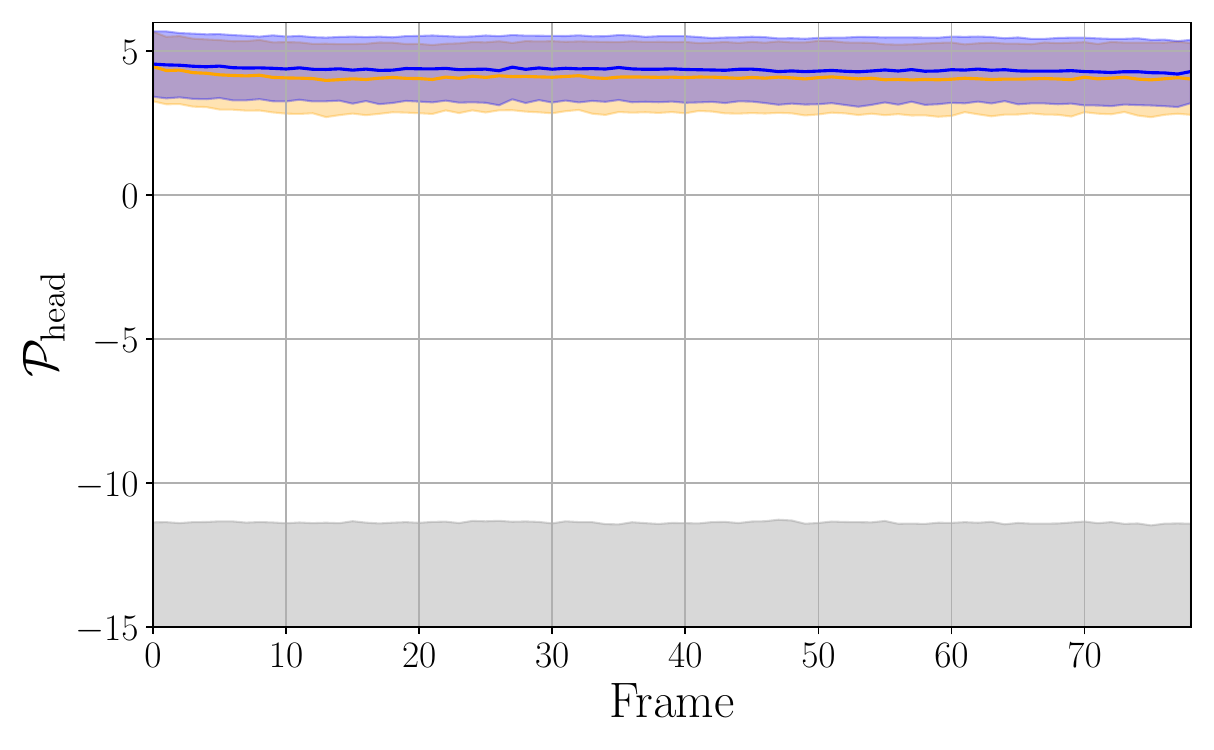}
    \caption{Dynamic}
    \label{fig:tempcons_Dynamic}
  \end{subfigure}
  \hfill
  \begin{subfigure}{0.32\textwidth}
    \centering
    \includegraphics[width=\linewidth]{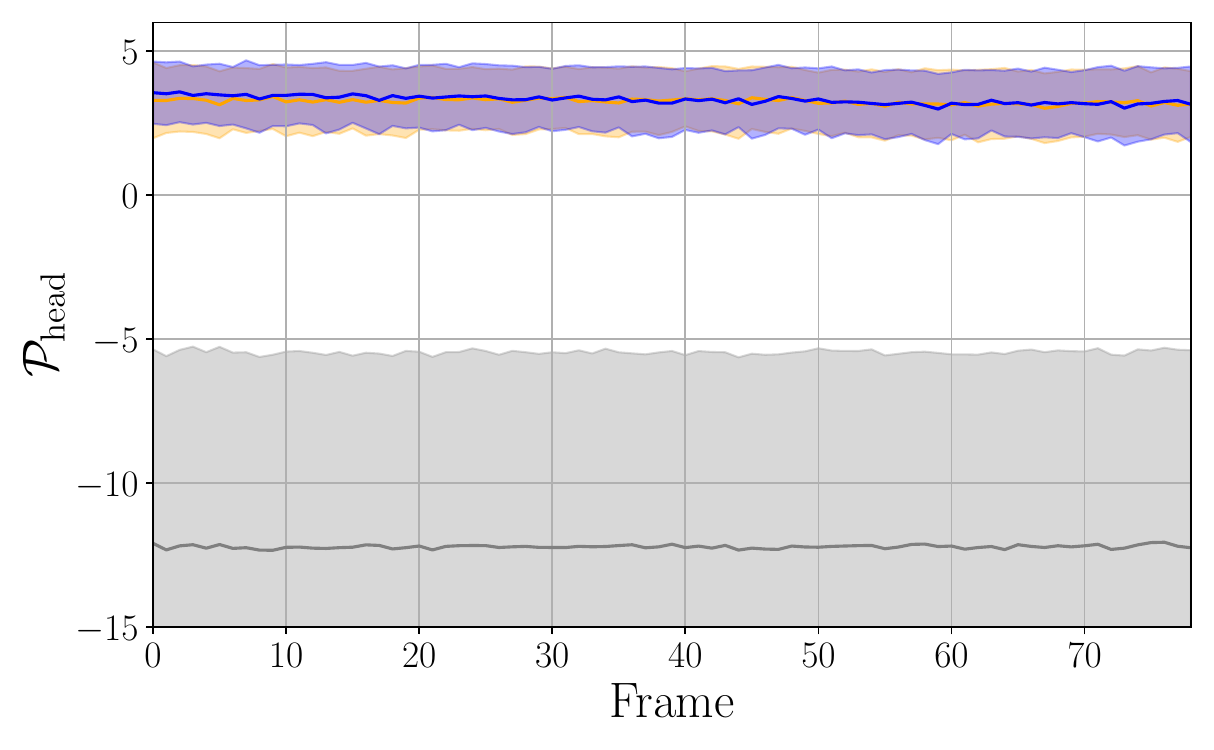}
    \caption{A4C}
    \label{fig:tempcons_a4c}
  \end{subfigure}
  \hfill
  \begin{subfigure}{0.32\textwidth}
    \centering
    \includegraphics[width=\linewidth]{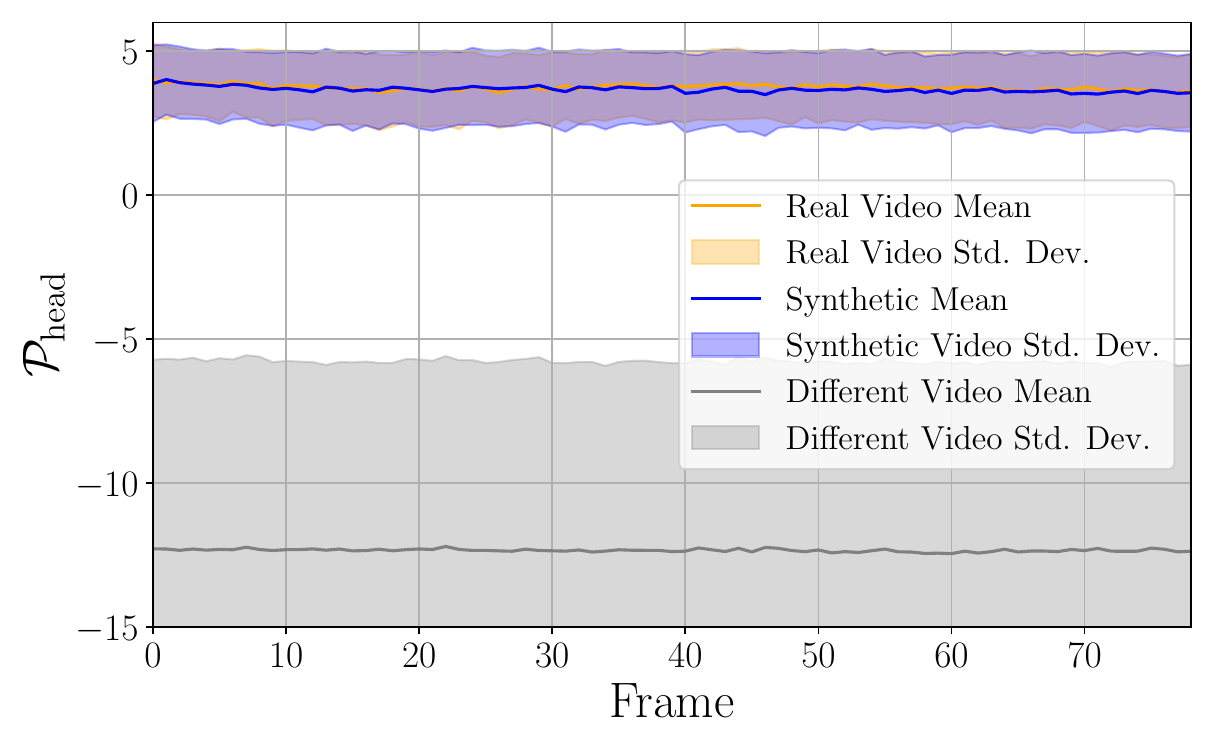}
    \caption{PSAX}
    \label{fig:tempcons_psax}
  \end{subfigure}
    \caption{Video consistency of real test videos in latent space. We compute the consistency of the first frame to the entire video of synthetic and real vidoes. As a comparison, we also do the same thing for a real frame with all frames from a different video. Overall, we can see that the model has consistently high score for frames belonging to the same video, proving that the model can be used to evaluate consistency.}
  \label{fig:tempcons}
\end{figure*}

Next, we test if we can use the models to evaluate temporal consistency. We do that by running the latent models on the real test videos and checking the anatomical distance between all the frames of any given video. We filter out all videos that are shorter than 80 frames, which is equal to \numprint{1258}, 210, and 297 videos for Dynamic, Pediatric (A4C), and Pediatric (PSAX) respectively. 
Additionally, we compare the consistency of one frame to a completely different video, which we expect to be low. 
Results are shown in Fig.~\ref{fig:tempcons}. As expected, the model  learns to pick up the consistency of the videos throughout the entire sequence. 
Importantly, the results show that there is no observable difference in terms of consistency between the real videos and the synthetic videos.  

\subsection{Downstream Evaluation}
\begin{table*}[!ht]
    \centering
    \resizebox{\textwidth}{!}{%
    \begin{tabular}{clcccccccccccccc}
    \toprule
        & & \multicolumn{4}{c}{\textbf{Dynamic}~\cite{ouyang2020video}} &&  \multicolumn{4}{c}{\textbf{Ped (A4C)}~\cite{reddy2023video}} && \multicolumn{4}{c}{\textbf{Ped (PSAX)}~\cite{reddy2023video}} \\
        & & {$R^2$}$\uparrow$ & {MAE}$\downarrow$ & {RMSE}$\downarrow$ & $N_{train}$ & & {$R^2$}$\uparrow$ & {MAE}$\downarrow$ & {RMSE}$\downarrow$ &$N_{train}$ && {$R^2$}$\uparrow$ & {MAE}$\downarrow$ & {RMSE}$\downarrow$ &$N_{train}$ \\
        \cmidrule{3-6}\cmidrule{8-11}\cmidrule{13-16}
        \multirow{2}{*}{\rotatebox{90}{Real}}& 
        Full \cite{ouyang2020video}          & 0.82 & 3.96 & 5.22 & 7456  && 0.74 & 4.21 & 5.88 & 2580  && 0.76 & 3.67 & 5.13 & 3559 \\
        & Subset                             & 0.70 & 4.98 & 6.65 & 1396  && 0.71 & 4.42 & 6.23 & 791   && 0.68 & 4.29 & 5.94 & 882 \\
        \cmidrule{3-6}\cmidrule{8-11}\cmidrule{13-16}
        \multirow{4}{*}{\rotatebox{90}{Synthetic}} &
         Full\cite{reynaud2024echonet} (repr.) 
                                           & 0.29 & 9.00 & 10.32 & 91399 && 0.57 & 5.45 & 7.59& 99362 && 0.06 & 8.74 & 10.15 & 97064\\
        & Random                           & 0.21 & 9.57 & 10.85 & 7456  && 0.52 & 6.17 & 8.01 & 2580  && 0.18 & 8.01 & 9.50& 3559\\
        & Recall small                     & -0.01&10.86 & 12.13& 1396  && 0.52 & 6.05 & 8.04  & 791   && -0.02 & 9.20 &10.61& 882 \\
        & Recall large                     & 0.28 & 9.12 & 10.38& 5808  && 0.56 & 5.79 & 7.68  & 3511  && 0.15 & 8.25 & 9.71 & 3812\\
        \bottomrule
    \end{tabular}
    }
    \caption{Performance of fully synthetic datasets on the downstream task of EF regression. We compare approaches using real data with approaches training exclusively on synthetic data. For the synthetic trainings we compare between sampling 100k videos ("Full"), sampling as many as we have real samples ("Subset") and recall informed approaches based on the results from \cref{sec:ModelRecall}.}
    \label{tab:downstream}
\end{table*}

Following previous work \cite{reynaud2023feature, reynaud2024echonet}, we estimate real-world performance by training a downstream model for EF estimation using various real and synthetic datasets \cite{ouyang2020video}. The dynamic models are trained from scratch, while pediatric models start from a model pre-trained on real dynamic data. 

We compare the performance gap between real and synthetic data in different scenarios. 
For the synthetic training sets we choose "Full" which consists of all privatised videos, "Random" which is a random subset equal to the size of the real dataset, and two recall-informed datasets. The first one trains on one synthetic video corresponding to each real videos (see \cref{tab:recallsummary} learned but not memorized), whereas the large one trains on five. 
To ensure a fair comparison, we limit the training on the full synthetic dataset to five epochs instead of 45.
Additionally, to mitigate conditioning effects of the LVDM, we run inference using an EF regressor pre-trained on real data to refine the ground-truth EF scores of synthetic videos. 
The results, shown in \cref{tab:downstream}, indicate that while the gap between real and synthetic data persists, there is a slight improvement when synthetic training videos are sampled based on our re-identification model. With only a fraction of the training samples, the "synaug" approaches reach equal performance. Despite that, training on the smallest synthetic datasets failed in all cases.

\section{Discussion}
The five percent threshold for determining memorization, based on \cite{dar2024unconditional}, was derived from a different dataset and should be re-evaluated for ultrasound videos. Additionally, as discussed in \cref{sec:exp_privacy}, we observe that the number of memorized samples is drastically lower for the same architecture compared to previous studies, as it depends on the trained privacy filter. This indicates that the method has potential for further improvements in robustness, as it may produce varied predictions for the same dataset. 
Finally, the downstream task still has not reached a performance level that is competitive with the real dataset. Reaching these levels requires additional research on the influence of recall and other advanced generative metrics. 

\section{Conclusion} 
In this paper, we apply privacy filtering techniques to quantitatively measure the quality, mode coverage, and real world utility of synthetic videos. 
Specifically, we show that applying privacy filters to the latent space of the generative model yields equal performance while improving the robustness when it comes to applying the same filters to other datasets. 
Furthermore, we show that the approach that is used to filter similar images, can be used to evaluate the latent space and faithfulness of the generated model and specifically the lack of coverage of the training distribution, which we show to be one of the reasons for limited performance of downstream tasks that are exclusively trained on synthetic data. 
In the future, we will work on analyzing and improving the robustness of the filter process and use our finding to increase the recall of unconditional generative models.

\noindent\textbf{Acknowledgements:} The authors gratefully acknowledge the scientific support and HPC resources provided by the Erlangen National High Performance Computing Center (NHR@FAU) of the Friedrich-Alexander-Universität Erlangen-Nürnberg (FAU) under the NHR projects b143dc and b180dc. NHR funding is provided by federal and Bavarian state authorities. NHR@FAU hardware is partially funded by the German Research Foundation (DFG) – 440719683. Support was also received from the ERC - project MIA-NORMAL 101083647 and DFG KA 5801/2-1, INST 90/1351-1. This work was supported by the UKRI Centre for Doctoral Training in Artificial Intelligence for Healthcare (EP/S023283/1) and Ultromics Ltd.
{\small
\bibliographystyle{ieee_fullname}
\bibliography{bib,main}
}

\end{document}